\documentclass[10pt,conference,letterpaper]{IEEEtran}

\usepackage{times,amsmath,epsfig, amssymb}

\usepackage[noend]{algpseudocode}
\algnewcommand\algorithmicforeach{\textbf{foreach}}
\algdef{S}[FOR]{Foreach}[1]{\algorithmicforeach\ #1\ \algorithmicdo}

\usepackage[]{algorithm}

\usepackage{mathtools}
\usepackage{esvect}

\usepackage[dvipsnames]{xcolor}
\usepackage{color, colortbl}
\definecolor{carminered}{rgb}{1.0, 0.0, 0.22}
\definecolor{whitesmoke}{rgb}{0.9, 0.9, 0.9}
\definecolor{graysmoke}{rgb}{0.8, 0.8, 0.8}
\definecolor{inviolet}{rgb}{0.6, 0, 0.7}
\definecolor{my_blue}{HTML}{90CAF9}
\definecolor{my_green}{HTML}{80CBC4}
\definecolor{my_amber}{HTML}{F48FB1}

\usepackage{soul}

\usepackage{comment}

\usepackage{setspace}
\usepackage{stfloats}

\usepackage{arydshln}
\usepackage{marvosym}

\usepackage{tabu}
\usepackage{multirow}
\usepackage{makecell}

\newcolumntype{C}[1]{>{\centering\arraybackslash}m{#1}}

\usepackage{hhline}

\usepackage{subfig}
\usepackage{adjustbox}
\usepackage{stackengine}

\definecolor{codegreen}{rgb}{0,0.6,0}
\definecolor{codegray}{rgb}{0.3,0.3,0.3}
\definecolor{codepurple}{rgb}{0.58,0,0.82}
\definecolor{backcolour}{rgb}{0.98,0.98,0.98}

\usepackage{csquotes}
\usepackage{hyperref}

\title{McDiarmid Drift Detection Methods \\ for Evolving Data Streams}

\author{
	{Ali Pesaranghader {\small $^{1}$ \textsuperscript{(\Letter)}}, Herna L. Viktor{\small $~^{1}$}, Eric Paquet{\small $~^{1,2}$} }%
	\vspace{1.6mm}\\
	\fontsize{10}{10}\selectfont\itshape
	$^{1}$\,School of Electrical Engineering and Computer Science,\\
	University of Ottawa, Ottawa, ON K1N 6N5, Canada\\
	\fontsize{9}{9}\selectfont\ttfamily\upshape
	\{apesaran, hviktor\}@uottawa.ca\\
	\vspace{1.2mm}\\
	\fontsize{10}{10}\selectfont\rmfamily\itshape
	$^{2}$\,National Research Council,\\
	1200 Montreal Roard, Ottawa, ON K1A 0R6, Canada\\
	\fontsize{9}{9}\selectfont\ttfamily\upshape
	eric.paquet@nrc-cnrc.gc.ca
}

\begin{document}
\maketitle

\begin{abstract}
Increasingly, Internet of Things (IoT) domains, such as sensor networks, smart cities, and social networks, generate vast amounts of data. Such data are not only unbounded and rapidly evolving. Rather, the content thereof dynamically evolves over time, often in unforeseen ways. These variations are due to so-called concept drifts, caused by changes in the underlying data generation mechanisms. In a classification setting, concept drift causes the previously learned models to become inaccurate, unsafe and even unusable. Accordingly, concept drifts need to be detected, and handled, as soon as possible. In medical applications and emergency response settings, for example, change in behaviours should be detected in near real-time, to avoid potential loss of life. To this end, we introduce the McDiarmid Drift Detection Method (MDDM), which utilizes McDiarmid's inequality \cite{mcdiarmid1989method} in order to detect concept drift. The MDDM approach proceeds by sliding a window over prediction results, and associate window entries with weights. Higher weights are assigned to the most recent entries, in order to emphasize their importance. As instances are processed, the detection algorithm compares a weighted mean of elements inside the sliding window with the maximum weighted mean observed so far. A significant difference between the two weighted means, upper-bounded by the McDiarmid inequality, implies a concept drift. Our extensive experimentation against synthetic and real-world data streams show that our novel method outperforms the state-of-the-art. Specifically, MDDM yields shorter detection delays as well as lower false negative rates, while maintaining high classification accuracies.
\end{abstract}


\section{Introduction}

Traditionally, machine learning algorithms assume that data are generated by a stationary distribution and collected prior to learning. These assumptions are not valid in evolving environments, where the underlying distributions may change over time: a phenomenon known as \textit{concept drift} \cite{gama2014survey, bifet2011data}. As a consequence, classification accuracy decreases as concept drifts take place. Therefore, adaptation to new distributions is essential to ensure the efficiency of the decision-making process.
An adaptive learning algorithm may utilize a drift detection method for detecting concept drifts in a data stream \cite{gama2014survey}.
Once the drift detector signals the presence of a concept drift, the learning algorithm updates its current model by considering the new distribution. 
For the learning process to be efficient, the drift detector must detect concept drifts rapidly, while maintaining low false negative and false positive rates.

In this paper, we introduce the McDiarmid Drift Detection Method (MDDM) which applies the McDiarmid inequality \cite{mcdiarmid1989method} and various weighting schemes in order to rapidly and efficiently detect concept drifts. Through numerous experiments, we show that MDDM finds abrupt and gradual concept drifts with shorter delays and with lower false negative rates, compared to the state-of-the-art.

This paper is organized as follows. \textit{Data stream classification} and \textit{concept drift} are formally defined in Sections \ref{sec_ds_classification} and \ref{sec_concept_drift_definition}, respectively. Section \ref{sec_adaptive_ds_learning} describes adaptive learning as a form of incremental learning from evolving data streams. Section \ref{sec_ddms} reviews the state-of-the-art for concept drift detection. In Section \ref{sec_fhddm_mddm}, we introduce the McDiarmid Drift Detection Methods (MDDMs). Next, in Section \ref{sec_experiments}, our approaches are compared with the start-of-the-art for both synthetic and real-world data streams. We conclude the paper and discuss future work in Section \ref{sec_conclusion}.


\section{Data Stream Classification}
\label{sec_ds_classification}

The primary objective of data stream classification is to build a model \textit{incrementally}, using the (current) available data, the so-called \textit{training data}, for predicting the label of unseen examples. Data stream classification may be defined as follows:

\begin{displayquote}
	Let a stream $S$ be a sequence of instances: $(\textbf{\textsc{x}}_1, y_1), (\textbf{\textsc{x}}_2, y_2), ..., (\textbf{\textsc{x}}_t, y_t)$ . The pair $(\textbf{\textsc{x}}_t, y_t)$ represents an instance arriving at time $t$, where $\textbf{\textsc{x}}_t$ is a vector containing $k$ attributes: $\textbf{\textsc{x}}_t = (x_1, x_2, ..., x_k)$, while $y_t$ is a \textit{class label} which belongs to a set of size $m$, $y_t \in \{c_1, c_2, ..., c_m\}$. Assume a target function $y_t = f(\textbf{\textsc{x}}_t)$ which maps an input vector to a particular class label. The learning task consists of incrementally building a model $\tilde{f}$ that approximates the function $f$ at all time. Naturally, an approximation which \textit{maximizes} the classification accuracy is preferred \cite{frias2015online}.
\end{displayquote}

Bifet et al.\ \cite{bifet2011data} recommend that incremental learning algorithms should fulfill four essential requirements for data stream mining: (1) the examples should be processed one-by-one and only once in the order of their arrival, (2) memory usage should be constrained as the size of a data stream is typically substantially larger than the size of the available memory, (3) all the calculations should be performed in real-time or at least, in near real-time, and (4) the outcome of the classification process should be available at any time.

\section{Concept Drift Definition}
\label{sec_concept_drift_definition}

The Bayesian Decision Theory is commonly employed in describing classification processes based on their prior probability distribution of classes, i.e.\ $p(y)$, and the class conditional probability distribution, i.e.\ $p(\textbf{\textsc{x}}|y)$ \cite{gama2014survey, vzliobaite2010learning}.
The classification decision is related to the posterior probabilities of the classes.
The posterior probability associated with class $c_i$, given instance $\textbf{\textsc{x}}$, is obtained by:

\begin{equation}
	\label{equ_nb_decision}
	p(c_i|\textbf{\textsc{x}}) = \frac{p(c_i) \cdot p(\textbf{\textsc{x}}|c_i)}{p(\textbf{\textsc{x}})}
\end{equation}

\noindent where $p(\textbf{\textsc{x}}) = \sum_{i=1}^{m}p(c_i) \cdot p(\textbf{\textsc{x}}|c_i)$ is the marginal probability distribution. Formally, if a concept drift occurs in between time $t_0$ and $t_1$ we have:

\begin{equation}
	\label{equ_joint_dist}
	\exists \textbf{\textsc{x}} : p_{t_0}(\textbf{\textsc{x}}, y) \ne p_{t_1}(\textbf{\textsc{x}},y)
\end{equation}

\noindent where $p_{t_0}$ and $p_{t_1}$ represent the joint probability distributions at time $t_0$ and $t_1$, respectively \cite{gama2014survey}.
Eq.\ \eqref{equ_joint_dist} implies that the data distribution at times $t_0$ and $t_1$ are distinct, as their joint probabilities differ. From Eq.\ \eqref{equ_nb_decision}, it may be observed that a concept drift may occur \cite{gama2014survey}:

\begin{itemize}
	\item As a result of a change in the prior probability distribution $p(y)$,
	\item As a result of a change in the class conditional probability distribution $p(\textbf{\textsc{x}}|y)$,
	\item As a result of a change in the posterior probability distribution $p(y|\textbf{\textsc{x}})$, thus affecting the classification decision boundaries.
\end{itemize}

Gama et al.\ \cite{gama2014survey}, {\v{Z}liobait{\.e} \cite{vzliobaite2010learning}, and Krempl et al.\ \cite{krempl2014open} classify changes into two types, namely \textit{real concept drift} and \textit{virtual concept drift}. A \textit{real concept drift} refers to the changes in $p(y|\textbf{\textsc{x}})$ which affects the decision boundaries or the target concept (as shown in Fig.\ \ref{fig_real_vs_virtual} (\hyperref[fig_real_vs_virtual_b]{b})).
On the other hand, \textit{virtual drift} is the result of a change in $p(\textbf{\textsc{x}})$, and subsequently in $p(\textbf{\textsc{x}}|y)$, but not in $p(y|\textbf{\textsc{x}})$. That is, a virtual drift is a change in the distribution of the incoming data which implies that the decision boundaries remain unaffected (as in Fig.\ \ref{fig_real_vs_virtual} (\hyperref[fig_real_vs_virtual_c]{c})). From a predictive perspective, adaptation is required once a real concept drift occurs, since the current decision boundary turns out to be obsolete \cite{gama2014survey, krempl2014open}.

\begin{figure}[h]
	\begin{center}
	\subfloat[Original Data\label{fig_real_vs_virtual_a}]{\includegraphics[scale=0.5]{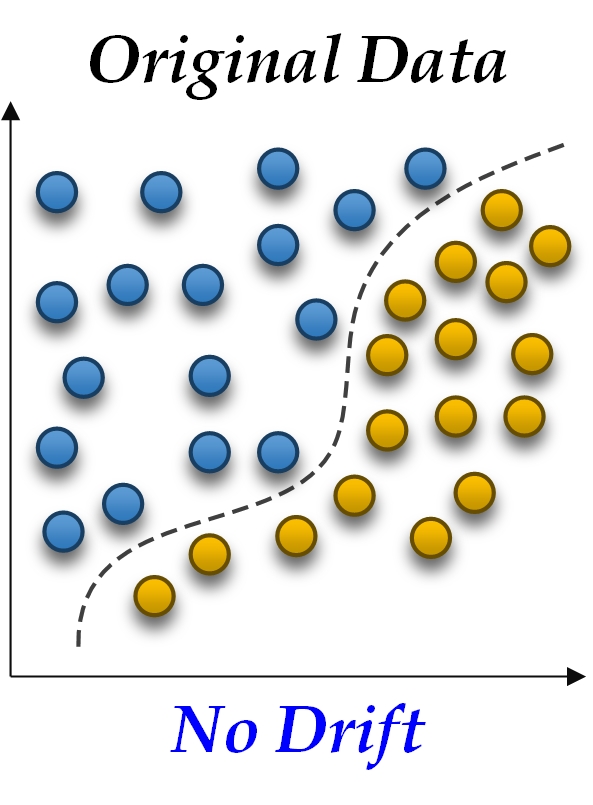}} \quad
	\subfloat[Real Drift\label{fig_real_vs_virtual_b}]{\includegraphics[scale=0.5]{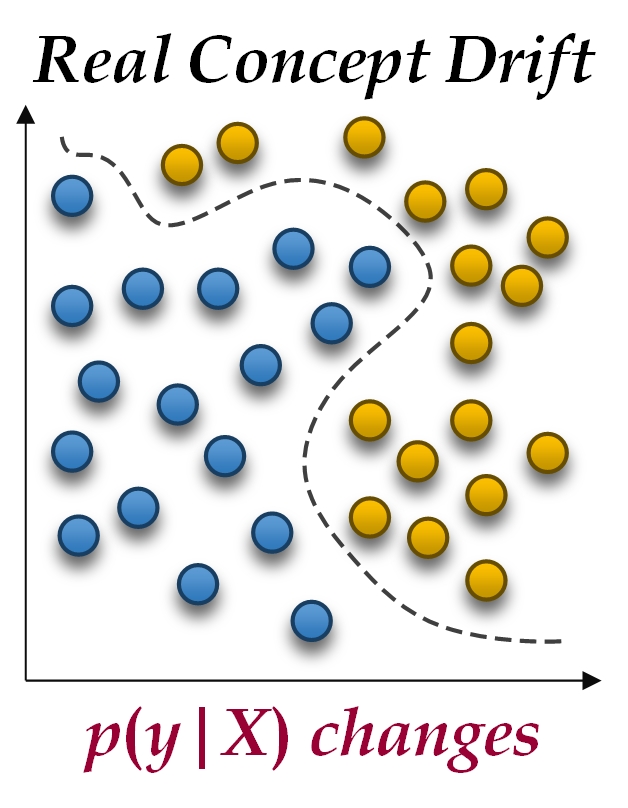}} \quad
	\subfloat[Virtual Drift\label{fig_real_vs_virtual_c}]{\includegraphics[scale=0.5]{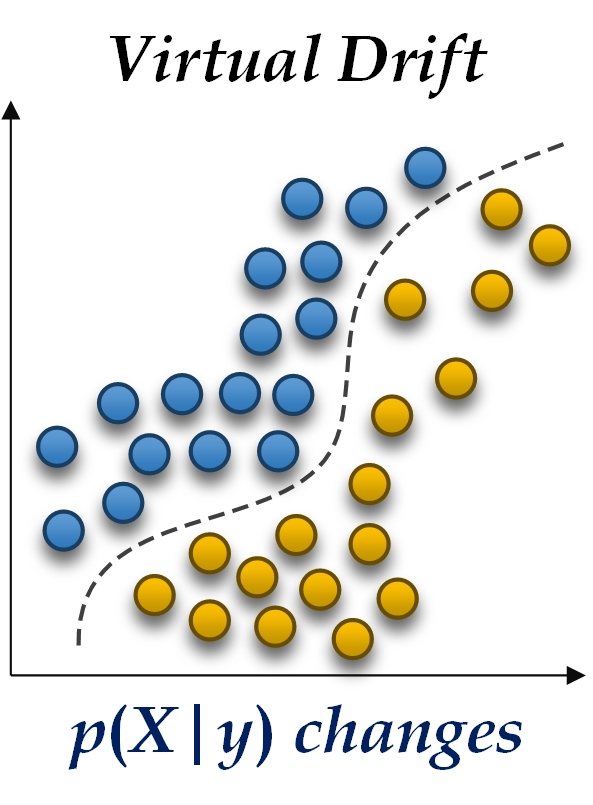}}
	\captionsetup{justification=centering}
	\caption{Real Concept Drift vs.\ Virtual Concept Drift \\ (Similar to Fig.\ 1 in \cite{gama2014survey})}
	\label{fig_real_vs_virtual}
	\end{center}
\end{figure}

\subsection{Concept Drift Patterns}
\label{subsec_concept_drift_pattern}

A concept drift may appear in different patterns \cite{vzliobaite2010learning}; as illustrated in Fig.\ \ref{fig_drift_patterns} (Note that colors represent different distributions). An \textit{abrupt} concept drift results from a sudden change in the data distribution.
On the other hand, a gradual concept drift results from a slow transition from one data distribution to the next. That is, the two patterns may coexist concurrently (Fig.\ \ref{fig_drift_patterns} (\hyperref[subfig_dp_b]{b})).
In an incremental concept drift, a sequence of data distributions appear during the transition. In re-occurring concept drift, a previously active concept reappears after some time, as shown in Fig.\ \ref{fig_drift_patterns} (\hyperref[subfig_dp_d]{d}). 
In practice, a mixture of different concept drifts may be present in the stream.

\begin{figure}[h]
	\centering \subfloat[Abrupt\label{subfig_dp_a}]{\includegraphics[scale=0.375]{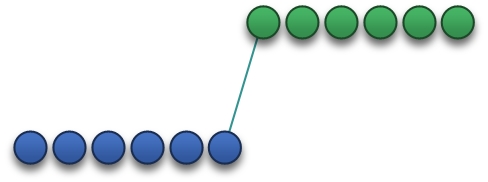}} \quad
	 \subfloat[Gradual\label{subfig_dp_b}]{\includegraphics[scale=0.375]{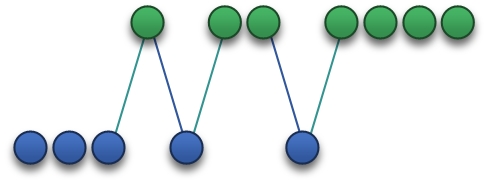}} \quad
	 \subfloat[Incremental\label{subfig_dp_c}]{\includegraphics[scale=0.375]{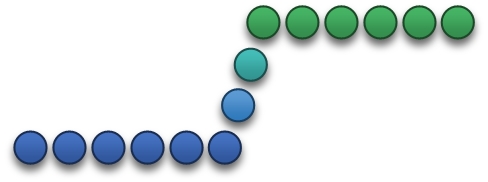}} \quad
	 \subfloat[Re-occurring\label{subfig_dp_d}]{\includegraphics[scale=0.375]{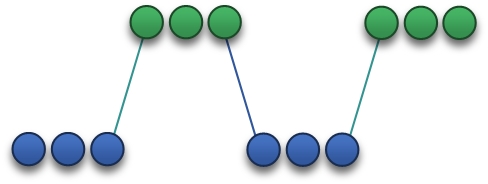}}
	 \caption{Patterns of Concept Drifts \\ (Similar to Fig.\ 2 in \cite{gama2014survey})}
	 \label{fig_drift_patterns}
\end{figure}


\section{Adaptive Data Stream Learning}
\label{sec_adaptive_ds_learning}

As learning algorithms are often trained in non-stationary environments, where concept drift is inevitable, they must have the capacity to adapt to new situations \cite{duda2017convergent}.
Gama et al.\ \cite{gama2014survey} defined adaptive learning as a form of advance incremental learning in which concept drifts are detected while the classification models are updated accordingly.
Adaptive learning algorithms must fulfill the following requirements in order to maintain high predictive performances \cite{pears2014detecting, pesaranghader2016fast, bifet2017classifier}: 
(1) \textit{Minimum false positive and false negative rates} -- an adaptive algorithm must detect concept drifts with a small number of false positives and false negatives. A high false positive rate involves more model retraining which in turn requires more computational resources \cite{vzliobaite2015towards}.
On the other hand, a high false negative rate reduces the classification accuracy, as the current model does not reflect the new distribution. 
(2) \textit{Short drift detection delay} -- An adaptive learning algorithm should detect concept drifts rapidly, and update its predictive model in quasi real-time in order to maintain the classification accuracy. 
(3) \textit{Robustness to noise} -- adaptive learners must be able to distinguish concept drift from noise. Indeed, no adaptation is required if noise is present in a stream.


\section{Concept Drift Detection Methods}
\label{sec_ddms}

Change detection methods refer to techniques and algorithms that detect concept drifts and distributional changes explicitly. Drift detection methods characterize and quantify concept drifts by discovering the change points or small time intervals during which concept drifts occur. Gama et al.\ \cite{gama2014survey} classify concept drift detectors into three groups:

\begin{enumerate}
	\item \textit{Sequential Analysis based Methods} sequentially evaluate prediction results as they become available. They alarm for concept drifts when a pre-defined threshold is met. The Cumulative Sum (CUSUM) and its variant PageHinkley (PH) \cite{page1954continuous} are representatives of this group.
	
	\item \textit{Statistical based Approaches} analyze statistical parameters such as the mean and the standard deviation associated with the predicted results in order to detect concept drifts. The Drift Detection Method (DDM) \cite{gama2004learning}, Early Drift Detection Method (EDDM) \cite{baena2006early}, Exponentially Weighted Moving Average (EWMA) \cite{ross2012exponentially}, and Reactive Drift Detection Method (RDDM) \cite{barros2017rddm} are members of this group. 
	
	\item \textit{Windows based Methods} usually utilize a fixed reference window for summarizing the past information and a sliding window for summarizing the most recent information. A significant difference in between the distributions of these two windows implies the occurrence of a drift. Statistical tests or mathematical inequalities, with the null-hypothesis indicating that the distributions are equal, are employed. The Adaptive Windowing (ADWIN) \cite{bifet2007learning}, the SeqDrift detectors \cite{pears2014detecting,sakthithasan2013one}, the Drift Detection Methods based on Hoeffding's Bound (HDDM\textsubscript{A-test} and HDDM\textsubscript{W-test}) \cite{frias2015online}, Fast Hoeffding Drift Detection Method (FHDDM) \cite{pesaranghader2016fast} and its stacking version (FHDDMS) \cite{pesaranghader2017reservoir} 
	are members of this family.
\end{enumerate}

CUSUM and its variant PageHinkley (PH) are some of the pioneer methods in the community. DDM, EDDM, and ADWIN have frequently been considered as benchmarks in the literature \cite{frias2015online, pesaranghader2016fast, baena2006early, bifet2007learning, huang2015drift}. RDDM, SeqDrift2, HDDMs, and FHDDM present similar performances. For these reasons, all these methods are evaluated in our experiments. Due to page limitations, we do not provide descriptions of these algorithms; therefore, we refer the interested reader to \cite{gama2014survey, pesaranghader2017reservoir} for further details. The pros and cons of these approaches are discussed below.

\textit{\textbf{Discussion --}} CUSUM and PageHinkley (PH) detect concept drifts from the deviation of the observed values from their mean and alarm for a drift when this difference exceeds a user-defined threshold. These algorithms are sensitive to the parameter values, resulting in a trade-off between false alarms and detecting true drifts \cite{gama2014survey,bifet2011data}.
DDM and EDDM require less memory as only a small number of variables is maintained \cite{gama2014survey}.
On the other hand, the ADWIN and SeqDrift2 approaches necessitate multiple subsets of the stream which lead to more memory consumption. They may computationally be expensive, due to the sub-window compression or reservoir sampling procedures.
Barros et al.\ \cite{barros2017rddm} observed that, RDDM leads to a higher classification accuracy compared to DDM, especially against datasets with gradual concept drift, despite an increase in false positives.
EDDM may frequently alarm for concept drift in the early stages of learning if the distances in between wrong predictions are small.  
HDDM and FHDDM employ the Hoeffding inequality \cite{hoeffding1963probability}. FHDDM differs from HDDM by sliding a window on prediction results for detecting concept drifts.
Recall that SeqDrift2 employs the Bernstein inequality \cite{bernstein1946theory} in order to detect concept drift. SeqDrift2 uses the sample variance, and assumes that the sampled data follow a normal distribution. This assumption may be too restrictive, in real-world domains. Further, the Bernstein inequality is conservative and requires a variance parameter, in contrast to, for instance, the Hoeffding inequality. These shortcomings may lead to a longer detection delay and a potential loss of accuracy. In the next section, we introduce McDiarmid Drift Detection Method (MDDM) for detecting concept drifts faster.


\section{McDiarmid Drift Detection Methods}
\label{sec_fhddm_mddm}

In a streaming environment, one may assume that old examples are either obsolete or outdated. Therefore, incremental learners should rely on the most recent examples for training, as the latter reflect the current situation more adequately.
Fading or weighting approaches are typically used by online learning algorithms to increase the weight attributed to the most recent instances \cite{gama2014survey}. 
This is important from an adaptive learning perspective, especially when a transition between two contexts is occurring. For instance, Klinkenberg \cite{klinkenberg2004learning} relies on an exponential weighting scheme $w_\lambda(\textsc{x}_i) = exp(-\lambda i)$, where $\lambda$ is a parameter and $i$ is the entry index, to assign lower weights to old examples.
Based on this observation, assigning higher weights to recent predictions could potentially result in a faster detection of concept drifts. In this section, we introduce the McDiarmid Drift Detection Methods (MDDMs) which utilizes a weighting scheme to ponderate the elements of its sliding window for faster detection of concept drifts. We also discuss variants of MDDM as well as the sensitivity of their parameters.

\subsection{McDiarmid Drift Detection Methods (MDDMs)}

The McDiarmid Drift Detection Method (MDDM) applies McDiarmid's inequality \cite{mcdiarmid1989method} to detect concept drifts in evolving data streams. The MDDM algorithm slides a window of size $n$ over the prediction results. It inserts a $1$ into the window if the prediction result is \textit{correct}; and $0$ otherwise. Each element in the window is associated with a weight, as illustrated in Fig.\ \ref{fig_mddm}, where $w_i < w_{i + 1}$. While inputs are processed, the weighted average of the elements of the sliding window is calculated, i.e.\ $\mu^t_w$, as well as the maximum weighted mean observed so far, i.e.\ $\mu^m_w$, as indicated in Eq.\ \eqref{equ_fhddm_1}.
\begin{equation}
	\label{equ_fhddm_1}
	\mbox{if} \quad \mu^m_w < \mu^t_w \Rightarrow \mu^m_w = \mu^t_w
\end{equation}

Recall that MDDM relies on the assumption that by weighting the prediction results associated with a sliding window, and by putting more emphasis on the most recent elements, concept drift could be detected faster and more efficiently. Given the rule $w_i < w_{i + 1}$, the elements located at the head of the window have higher weights than those located at the tail.
Different weighting schemes have been considered including the \textit{arithmetic} and the \textit{geometric} schemes.
The arithmetic scheme is given by $w_i = 1 + (i - 1) * d$, where $d \ge 0$ is the difference between two consecutive weights. The geometric scheme is given by $w_i = r^{(i - 1)}$, where $r \ge 1$ is the ratio of two consecutive weights. In addition, we employ the Euler scheme which is defined by $r = e^{\lambda}$ where $\lambda \ge 0$.
We have implemented three weighted drift detection methods based on these three schemes: MDDM-A (A for arithmetic), MDDM-G (G for geometric), and MDDM-E (E for Euler)\footnote{The source codes are available at \url{https://www.github.com/alipsgh/} (One may use them with the MOA framework \cite{bifet2010moa}).}.
All these methods are described below. As the prediction results are processed one-by-one, the algorithm calculates the weighted average of the elements inside the sliding window, and simultaneously updates two variables $\mu^t_w$ (i.e.\ the current weighted average) and $\mu^m_w$ (i.e.\ the maximum weighted average observed so far). A significant difference between $\mu^m_w$ and $\mu^t_w$ implies a concept drift.

\begin{figure}[h]
	\begin{center}
		\includegraphics[scale=0.45]{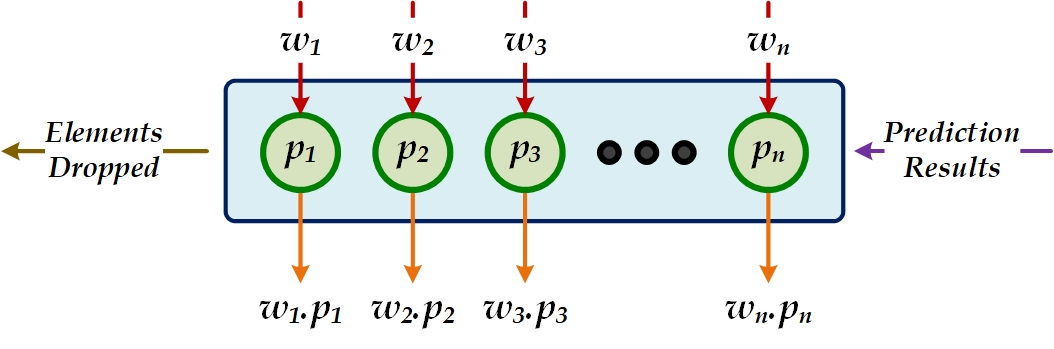}
		\caption{McDiarmid Drift Detection Method (General Scheme)}
		\label{fig_mddm}
	\end{center}
\end{figure}

On the basis of the PAC learning model \cite{mitchell1997machine}, the accuracy should increase or stay constant as the number of instances increases; otherwise, the possibility of facing drifts increases \cite{gama2004learning, sebastiaofading}.
Thus, the value of $\mu^m_w$ should increase or remain constant as more instances are processed. In other words, the possibility of facing a concept drift increases if $\mu^m_w$ does not change and $\mu^t_w$ decreases over time. Finally, as shown by Eq.\ \eqref{equ_fhddm_2}, a significant difference in between $\mu^m_w$ and $\mu^t_w$ indicates the occurrence of a drift:
\begin{equation}
\label{equ_fhddm_2}
\Delta \mu = \mu^m_w - \mu^t_w \geq \varepsilon_d \Rightarrow \mbox{Drift := True}
\end{equation}

The McDiarmid inequality \cite{mcdiarmid1989method} is employed to determine if the difference is deemed significant. \newline

\noindent \textbf{Theorem I: McDiarmid's Inequality \label{mcdiarmid_theory}--} Let $X_1, X_2, ..., X_n$ be $n$ independent random variables all taking values in the set $\chi$. Further, let $f:\chi^n \mapsto \mathbb{R}$ be a function of $X_1, ..., X_n$ that satisfies $\forall i, \forall x_1, ..., x_n, x_i' \in \chi$, $$|f\left(x_1,...,x_i,...,x_n\right) - f\left(x_1,...,x_i',...,x_n\right)| \le c_i.$$ This implies that replacing the $i^{th}$ coordinate $x_{i}$ by some arbitrary value changes the function $f$ by at most $c_{i}$. Then, for all $\varepsilon_M > 0$, we have:
\begin{equation}
	\label{equ_MDDM_test_1}
	\Pr\{E[f] - f \ge \varepsilon_M\} \le \exp\left(-\frac{2\varepsilon_M^2}{\sum_{i=1}^{n}c_i^2}\right)
\end{equation}

Consequently, for a given confidence level $\delta_M$, the value of $\varepsilon_M$ is obtained by:
\begin{equation}
	\label{equ_MDDM_test_2}
	\varepsilon_M = \sqrt{\frac{\sum_{i=1}^{n}c_i^2}{2}\ln\frac{1}{\delta_M}}
\end{equation}

\noindent \textbf{Corollary I: MDDM test \label{MDDM_test}--} In a streaming context, assume $\mu^t_w$ is the weighted mean of a sequence of $n$ random entries, at time $t$, and $\mu^m_w$ is the maximum weighted mean observed so far. Recall that each entry $p_i$ is in $\{0, 1\}$ and has a weight of $w_i$. Let $\Delta \mu_w = \mu^m_w - \mu^t_w \ge 0$ be the difference between these two weighted means. Given the confidence level $\delta_w$, the McDiarmid inequality detects a drift if $\Delta \mu_w \ge \varepsilon_w$, where
\begin{equation}
	\label{equ_MDDM_test_3}
	\varepsilon_w = \sqrt{\frac{\sum_{i=1}^{n}v_i^2}{2}\ln\frac{1}{\delta_w}}
\end{equation}

\noindent and where $v_i$ is given by
\begin{equation}
	\label{equ_MDDM_test_4}
	v_i = \frac{w_i}{\sum_{i=1}^{n}w_i}
\end{equation}

All three of our MDDM approaches apply Corollary \hyperref[MDDM_test]{I} in order to detect concept drift. In addition, the McDiarmid inequality allows for comparing the expectation of a function of the random variables, such as the maximum, with the function ``per se''. This stands in contrast with, for instance the Hoeffding inequality \cite{hoeffding1963probability}, in which the comparison is restricted to the expectation of the random variables with their empirical mean.

The pseudocode for the MDDM algorithm appears in Algorithm \ref{algo_mddm}. Firstly, the \textsc{Initialize} function initializes the parameters, including the window size $n$, confidence level $\delta_w$, $\varepsilon_w$, the sliding window $win$, and $\mu^m_w$. As the data stream examples are processed, the prediction results are pushed into the window (lines 8-14). The algorithm updates the variables $\mu^t_w$ and $\mu^m_w$ over time (lines 15-17). Finally, a drift is detected if $(\mu_w^m - \mu_w^t) \ge \varepsilon_w$ (lines 18-21). Recall that we have $w_i = 1 + (i - 1) * d$ for MDDM-A, $w_i = r^{(i - 1)}$ for MDDM-G, and $w_i = e^{\lambda(i - 1)}$ for MDDM-E.

\begin{algorithm}[h]
	\caption{McDiarmid Drift Detection Method}
	\label{algo_mddm}
	\small
	\begin{algorithmic}[1]
		
		\Function{Initialize}{$windowSize, delta$}
		\State $(n, \delta_w) \leftarrow (windowSize, delta)$
		\State $\varepsilon_w$ = \Call{CalculateEpsilon}{}()
		\State \Call{Reset}{}()
		\EndFunction
		
		\Function{Reset}{}()
		\State $\mbox{win} = []$ \Comment{Creating an empty sliding window.}
		\State $\mu^m_w = 0$
		\EndFunction
		
		\Function{Detect}{$pr$} \Comment{The \emph{pr} is 1 for correct predictions, 0 otherwise.}
		\If{$\mbox{win}.size() = n$}
		\State $\mbox{win}.tail.drop()$ \Comment{Dropping an element from the tail.}
		\EndIf
		\State $\mbox{win}.push(pr)$ \Comment{Pushing an element into the head.}
		\If{$\mbox{win}.size() < n$}
		\State \Return{False}
		\Else
		\State $\mu^t_w$ = \Call{GetWeightedMean}{}()
		\If{$\mu^m_w < \mu^t_w$}
		\State $\mu^m_w = \mu^t_w$
		\EndIf
		\State $\Delta \mu_w = \mu^m_w - \mu^t_w$
		\If {$\Delta \mu_w \geq \varepsilon_w$}
		\State \Call{Reset}{}() \Comment{Resetting parameters.}
		\State \Return{True} \Comment{Signaling for an alarm.}
		\Else
		\State \Return{False}
		\EndIf
		\EndIf
		\EndFunction
		
		\Function{CalculateEpsilon}{}()
			\State $S = \sum_{i=1}^{n}v_i^2$ \Comment{$v_i = w_i/\sum_{i=1}^{n}w_i$}
			\State \Return{$\sqrt{\frac{S}{2} \ln\frac{1}{\delta_w}}$} 
		\EndFunction
		
		\Function{GetWeightedMean}{}()
			\State \Return{$\sum_{i=1}^{n}(p_i \times w_i) / \sum_{i=1}^{n} w_i$}
		\EndFunction
		
	\end{algorithmic}
\end{algorithm}

\subsection{Discussion On Variants of MDDM}

Recall that the MDDM-A approach employs an arithmetic scheme $w_i = 1 + (i - 1) * d$, where $w_{i+1} - w_i = d$ meaning that the weights increase \textit{linearly}. On the the other hand, MDDM-G applies the geometric scheme $w_i = r^{(i - 1)}$, indicating that the weights increase \textit{exponentially} with $w_{i+1}/w_{i} = r$ (Note that $r=e^\lambda$ for MDDM-E). The linear or exponential nature of the weighting scheme affects the detection delay and the false positive rate. That is, the exponential weighting scheme often results in a faster concept drift detection, but at the expense of a higher false positive rate if compared to the linear weighting scheme. These statements are supported by experimental results in Section \ref{sec_experiments}.

\subsection{Parameters Sensitivity Analysis}

The parameters $n$ and $\delta_w$ are inversely proportional with respect to $\varepsilon_w$. That is, as the value of $n$ increases, the value of $\varepsilon_w$ decreases. This implies that, as more observations become available, a more optimistic error bound should be applied. On the other hand, as the value of $\delta_w$ decreases, the values of $\varepsilon_w$ increases (i.e.\ the bound becomes more conservative).
The parameter $d$ in MDDM-A controls the scale of the weights assigned to the sliding window elements. The value of $\varepsilon_w$ increases, as the value of $d$ increases. Larger values of $d$ lead to faster drift detection, since higher weights are assigned to the element located at the head of the window; however, the false positive rate may increase.
MDDM-G and MDDM-E behave similarly; the scale of their weight is determined by their parameters $r$ and $\lambda$, respectively. That is, a higher $r$ or $\lambda$ leads to a shorter detection delay, but at the expense of a higher false positive rate. In order to set the default values of these parameters, we conducted a number of experiments against various synthetic data streams. We gradually increased the values of these parameters to find the optimal values: $\delta_w = 10^{-6}$, $d = 0.01$, $r = 1.01$, and $\lambda = 0.01$.


\section{Experimental Evaluation}
\label{sec_experiments}

\subsection{Benchmarking Data Streams}
\label{subsec_data_streams}

\subsubsection{Synthetic Data Streams}

We generated four synthetic data streams from \textsc{Sine1}, \textsc{Mixed}, \textsc{Circles} and \textsc{LED}, which are all widely found in the literature \cite{frias2015online, pesaranghader2016fast, gama2004learning, ross2012exponentially, barros2017rddm, pesaranghader2016framework, bifet2009new, de2017wilcoxon}. Each data stream consists of $100,000$ instances.
A class noise of 10\% was added to each stream in order to evaluate the robustness of the drift detectors against noisy data\footnote{Available at: \url{https://www.github.com/alipsgh/data_streams/}.}. The synthetic data streams are described below.

\begin{itemize}
	\item \textsc{\textbf{Sine1}} $\cdot$ \textit{with abrupt drift}: It has two attributes $x$ and $y$ uniformly distributed on the interval $[0, 1]$. The classification function is $y = sin(x)$. Instances are classified as positive if they are under the curve; otherwise, they are negative. At a drift point, the classification is reversed \cite{pesaranghader2016fast,gama2004learning,baena2006early, ross2012exponentially}.
	\item \textsc{\textbf{Mixed}} $\cdot$ \textit{with abrupt drift}: The dataset has two numeric attributes $x$ and $y$ distributed in $[0, 1]$ with two boolean attributes $v$ and $w$. The instances are classified as positive if at least two of the three following conditions are satisfied: $v, w, y < 0.5 + 0.3 * sin(3\pi x)$. The classification is reversed when drift points occur \cite{pesaranghader2016fast,gama2004learning,barros2017rddm}.
	\item \textsc{\textbf{Circles}} $\cdot$ \textit{with gradual drift}: 
	It has two attributes $x$ and $y$ distributed in $[0, 1]$. The classification function is the circle equation $(x - x_c)^2 + (y - y_c)^2 = r_c^2$ where $(x_c, y_c)$ and $r_c$ are the center and the radius of the circle, respectively. Instances inside the circle are classified as positive.
	Four different circles are employed in order to simulate concept drift \cite{pesaranghader2016fast,gama2004learning,baena2006early}.
	\item \textsc{\textbf{LED}} $\cdot$ \textit{with gradual drift}: The objective of this dataset is to predict the digit on a seven-segment display, where each digit has a 10\% chance of being displayed. The dataset has 7 class attributes, and 17 irrelevant ones. Concept drift is simulated by interchanging relevant attributes \cite{frias2015online, bifet2009new,de2017wilcoxon}.
\end{itemize} 

\textit{Concept Drift Simulation --} Following Bifet et al.\ \cite{bifet2009new}, we used the sigmoid function to simulate \textit{abrupt} and \textit{gradual} concept drifts. The function determines the probability of belonging to a new context during a transition between two concepts. The transition length $\zeta$ allows 
to simulate abrupt or gradual concept drifts. The value was set to $50$ for abrupt concept drifts, and to $500$ for gradual concept drifts in all our experiments. To summarize, the drifts occur at every $20,000$ instances in \textsc{Sine1} and \textsc{Mixed} with $\zeta = 50$ for \textit{abrupt drift}, and at every $25,000$ instances in \textsc{Circles} and \textsc{LED} with $\zeta = 500$ for \textit{gradual drift}.

\subsubsection{Real-world Data Streams}

We extended our experiments to real-world data streams\footnote{Available at: \url{https://moa.cms.waikato.ac.nz/datasets/2013/}.}; which are frequently employed in the online learning and adaptive learning literature \cite{frias2015online,pesaranghader2016fast,gama2004learning,baena2006early,ross2012exponentially,pesaranghader2016framework,bifet2009new,escovedo2018detecta}. Three data streams were selected in our comparative study.
\begin{itemize}
	\item \textsc{\textbf{Electricity}} $\cdot$ It contains $45,312$ instances, with $8$ input attributes, recorded every half hour for two years by the Australian New South Wales Electricity. The classifier must predict a rise (\textit{Up}) or a fall (\textit{Down}) in the electricity price. The concept drift may result from changes in consumption habits or unexpected events \cite{zliobaite2013good}.
	
	\item \textsc{\textbf{Forest Covertype}} $\cdot$ It consists of $54$ attributes with $581,012$ instances describing $7$ forest cover types for $30 \times 30$ meter cells obtained from US Forest Service (USFS) information system, for $4$ wilderness areas located in the Roosevelt National Forest of Northern Colorado \cite{blackard1999comparative}.
	
	\item \textsc{\textbf{Poker hand}} $\cdot$ It is composed of $1,000,000$ instances, where each instance is an example of five cards drawn from a standard $52$ cards deck. Each card is described by two attributes (suit and rank), for a total of ten predictive attributes. The classifier predicts the poker hand \cite{olorunnimbe2017dynamic}.
\end{itemize}

\subsection{Experiment Settings}

We used the MOA framework \cite{bifet2010moa} for our experiments. We selected Hoeffding Tree (HT) \cite{domingos2000mining} and Naive Bayes (NB) as our incremental classifiers; and compared MDDMs with CUSUM, PageHinkley, DDM, EDDM, RDDM, ADWIN, SeqDrift2, HDDMs, and FHDDM. The default parameters were employed for both the classifiers and the drift detection methods. The algorithms were evaluated \textit{prequentially} which means that an instance is first tested and then used for training \cite{krawczyk2017ensemble}.

Pesaranghader et al.\ \cite{pesaranghader2016fast} introduced the \textit{acceptable delay length} notion for measuring detection delay and for determining true positive (TP), false positive (FP), and false negative (FN) rates. The acceptable delay length $\Delta$ is a threshold that determines how far a given alarm should be from the true location of a concept drift to be considered a true positive.
That is, we maintain three variables to count the numbers of true positives, false negatives and false positives which are initially set to zero.
Therefore, the number of true positives is incremented when the drift detector alarm occurs within the acceptable delay range. Otherwise, the number of false negatives is incremented as the alarm occurred too late. In addition, the false positive value is incremented when a false alarm occurs outside the acceptable delay range. Following this approach, we set $\Delta$ to $250$ for the \textsc{Sine1}, \textsc{Mixed}, and to $1000$ for the \textsc{Circles} and \textsc{LED} data streams. A longer $\Delta$ should be considered for data streams with gradual drifts in order to avoid a false negative increase.

Following \cite{pesaranghader2016fast}, for both MDDMs and FHDDM, the window size was set to $25$ for the \textsc{Sine1} and \textsc{Mixed}, and to $100$ for the \textsc{Circles} and \textsc{LED} data streams. We used a wider window for the \textsc{Circles} and \textsc{LED} data streams in order to better detect gradual concept drifts. These window sizes were chosen in order to have shorter detection delay, as well as lower false positive and false negative rates. Experiments were performed on an Intel Core i$5$ @ $2.8$ GHz with $16$ GB of RAM running Apple OS X Yosemite.

\subsection{Experiments and Discussion}

\subsubsection{Synthetic Data Streams}

Our experimental results against the synthetic data streams are presented in Tables \ref{tab_exp_syn_1} and \ref{tab_exp_syn_2}. Recall that as we are aware of the locations of drifts in synthetic data streams, we can evaluate the detection delay, true positive (TP), false positive (FP) and false negative (FN). We discuss the experimental results in the following:

\begin{table*}[ht]

	\begin{center}
		
		\caption{\small Hoeffding Tree and Naive Bayes with Drift Detectors against Synthetic Data Streams with Abrupt Change ($\zeta = 50$)}
		\label{tab_exp_syn_1}
		\def\arraystretch{1.01}
		\setlength\tabcolsep{1.5pt}
		\scriptsize
		
		\begin{tabu}{cr|c|c|c|c|c||c|c|c|c|c|}
			\cline{3-12}
			&             & \multicolumn{5}{c||}{Hoeffding Tree (HT)}    & \multicolumn{5}{c|}{Naive Bayes (NB)}    \\ \cline{2-12}
			\multicolumn{1}{c|}{}                         & Detector    & Delay & TP & FP & FN & Accuracy & Delay & TP & FP & FN & Accuracy \\ \hline
			
			
			\multicolumn{1}{||c||}{\multirow{13}{*}{\rotatebox[]{90}{\textsc{\textbf{Sine1}} - Abrupt}}}
			
			& \cellcolor{my_blue}MDDM-A
				& \cellcolor{my_blue}\textbf{38.60 $\pm$ 3.38} & \cellcolor{my_blue}4.00 & \cellcolor{my_blue}\textbf{0.21 $\pm$ 0.43} & \cellcolor{my_blue}0.00 & \cellcolor{my_blue}\textbf{87.07 $\pm$ 0.16} 
				& \cellcolor{my_blue}\textbf{38.55 $\pm$ 3.35} & \cellcolor{my_blue}4.00 & \cellcolor{my_blue}\textbf{0.13 $\pm$ 0.34} & \cellcolor{my_blue}0.00 & \cellcolor{my_blue}\textbf{86.08 $\pm$ 0.25} \\
			\multicolumn{1}{||c||}{}                        
			& \cellcolor{my_blue}MDDM-G
				& \cellcolor{my_blue}\textbf{38.56 $\pm$ 3.36} & \cellcolor{my_blue}4.00 & \cellcolor{my_blue}\textbf{0.20 $\pm$ 0.42} & \cellcolor{my_blue}0.00 & \cellcolor{my_blue}\textbf{87.07 $\pm$ 0.16}
				& \cellcolor{my_blue}\textbf{38.47 $\pm$ 3.35} & \cellcolor{my_blue}4.00 & \cellcolor{my_blue}\textbf{0.14 $\pm$ 0.35} & \cellcolor{my_blue}0.00 & \cellcolor{my_blue}\textbf{86.08 $\pm$ 0.25} \\
			\multicolumn{1}{||c||}{}                      
			& \cellcolor{my_blue}MDDM-E
				& \cellcolor{my_blue}\textbf{38.56 $\pm$ 3.36} & \cellcolor{my_blue}4.00 & \cellcolor{my_blue}\textbf{0.20 $\pm$ 0.42} & \cellcolor{my_blue}0.00 & \cellcolor{my_blue}\textbf{87.07 $\pm$ 0.16}
				& \cellcolor{my_blue}\textbf{38.46 $\pm$ 3.35} & \cellcolor{my_blue}4.00 & \cellcolor{my_blue}\textbf{0.14 $\pm$ 0.35} & \cellcolor{my_blue}0.00 & \cellcolor{my_blue}\textbf{86.08 $\pm$ 0.25} \\ \cline{2-12} 
			\multicolumn{1}{||c||}{}
			& CUSUM       
				& 86.89 $\pm$ 4.47 & 4.00 & \textbf{0.24 $\pm$ 0.47} & 0.00 & 86.94 $\pm$ 0.15
				& 83.27 $\pm$ 6.96 & 3.99 $\pm$ 0.10 & 0.71 $\pm$ 0.86 & 0.01 $\pm$ 0.10 & 85.96 $\pm$ 0.25 \\
			\multicolumn{1}{||c||}{}
			& PageHinkley 
				& 229.24 $\pm$ 13.20 & 2.30 $\pm$ 1.07 & 1.71 $\pm$ 1.08 & 1.70 $\pm$ 1.07 & 86.06 $\pm$ 1.34
				& 175.07 $\pm$ 24.72 & 3.71 $\pm$ 0.50 & 0.35 $\pm$ 0.54 & 0.29 $\pm$ 0.50 & 85.69 $\pm$ 0.27 \\
			\multicolumn{1}{||c||}{}
			& DDM         
				& 163.11 $\pm$ 22.73 & 3.36 $\pm$ 0.77 & 3.30 $\pm$ 2.20 & 0.64 $\pm$ 0.77 & 86.06 $\pm$ 1.34
				& 179.18 $\pm$ 26.83 & 2.87 $\pm$ 0.84 & 3.09 $\pm$ 1.88 & 1.13 $\pm$ 0.84 & 82.39 $\pm$ 4.32 \\
			\multicolumn{1}{||c||}{}
			& EDDM        
				& 243.83 $\pm$ 14.25 & 0.22 $\pm$ 0.44 & 33.90 $\pm$ 11.61 & 3.78 $\pm$ 0.44 & 84.71 $\pm$ 0.55
				& 234.28 $\pm$ 22.22 & 0.57 $\pm$ 0.64 & 33.53 $\pm$ 11.50 & 3.43 $\pm$ 0.64 & 83.44 $\pm$ 2.87 \\
			\multicolumn{1}{||c||}{}
			& RDDM        
				& 93.63 $\pm$ 7.57 & 4.00 & 4.72 $\pm$ 3.58 & 0.00 & 86.79 $\pm$ 0.18
				& 89.72 $\pm$ 16.45 & 3.99 $\pm$ 0.10 & 3.93 $\pm$ 2.91 & 0.01 $\pm$ 0.10 & 85.98 $\pm$ 0.27 \\ 
			\multicolumn{1}{||c||}{}
			& ADWIN       
				& 63.84 $\pm$ 1.12 & 4.00 $\pm$ 0.00 & 7.31 $\pm$ 3.18 & 0.00 & 86.67 $\pm$ 0.21
				& 63.92 $\pm$ 0.80 & 4.00 $\pm$ 0.00 & 3.86 $\pm$ 1.09 & 0.00 & 85.93 $\pm$ 0.23 \\ 
			\multicolumn{1}{||c||}{}
			& SeqDrift2
				& 200.00 & 4.00 & 4.83 $\pm$ 1.16 & 0.00 & 86.53 $\pm$ 0.15
				& 200.00 & 4.00 & 4.26 $\pm$ 0.58 & 0.00 & 85.59 $\pm$ 0.25 \\
			\multicolumn{1}{||c||}{}
			& HDDMA-test
				& 57.62 $\pm$ 11.81 & 4.00 & 0.71 $\pm$ 0.89 & 0.00 & \textbf{87.01 $\pm$ 0.16}
				& 88.03 $\pm$ 25.73 & 3.97 $\pm$ 0.17 & 0.35 $\pm$ 0.55 & 0.03 $\pm$ 0.17 & 85.95 $\pm$ 0.25 \\
			\multicolumn{1}{||c||}{}
			& HDDMW-test
				& \textbf{35.70 $\pm$ 2.95} & 4.00 & 0.46 $\pm$ 0.68 & 0.00 & \textbf{87.07 $\pm$ 0.15}
				& \textbf{35.52 $\pm$ 3.10} & 4.00 & 0.41 $\pm$ 0.58 & 0.00 & \textbf{86.09 $\pm$ 0.25} \\
			\multicolumn{1}{||c||}{}
			& FHDDM
				& 40.65 $\pm$ 3.15 & 4.00 & \textbf{0.10 $\pm$ 0.33} & 0.00 & \textbf{87.07 $\pm$ 0.16}
				& 40.48 $\pm$ 3.37 & 4.00 & \textbf{0.04 $\pm$ 0.20} & 0.00 & \textbf{86.08 $\pm$ 0.25} \\ 
			
			\hline								
			\multicolumn{9}{c}{} \\ [-0.6em]
			\hline
			
			
			\multicolumn{1}{||c||}{\multirow{13}{*}{\rotatebox[]{90}{\textsc{\textbf{Mixed}} - Abrupt}}}
			
			& \cellcolor{my_blue}MDDM-A      
				& \cellcolor{my_blue}\textbf{38.38 $\pm$ 3.66} & \cellcolor{my_blue}4.00 & \cellcolor{my_blue}\textbf{1.11 $\pm$ 1.15} & \cellcolor{my_blue}0.00 & \cellcolor{my_blue}\textbf{83.36 $\pm$ 0.11}
				& \cellcolor{my_blue}\textbf{38.52 $\pm$ 3.81} & \cellcolor{my_blue}4.00 & \cellcolor{my_blue}\textbf{0.69 $\pm$ 0.89} & \cellcolor{my_blue}0.00 & \cellcolor{my_blue}\textbf{83.37 $\pm$ 0.09} \\
			\multicolumn{1}{||c||}{}
			& \cellcolor{my_blue}MDDM-G      
				& \cellcolor{my_blue}\textbf{38.28 $\pm$ 3.64} & \cellcolor{my_blue}4.00 & \cellcolor{my_blue}\textbf{1.19 $\pm$ 1.21} & \cellcolor{my_blue}0.00 & \cellcolor{my_blue}\textbf{83.36 $\pm$ 0.11}
				& \cellcolor{my_blue}\textbf{38.41 $\pm$ 3.81} & \cellcolor{my_blue}4.00 & \cellcolor{my_blue}\textbf{0.70 $\pm$ 0.89} & \cellcolor{my_blue}0.00 & \cellcolor{my_blue}\textbf{83.37 $\pm$ 0.09} \\ 
			\multicolumn{1}{||c||}{}
			& \cellcolor{my_blue}MDDM-E      
				& \cellcolor{my_blue}\textbf{38.28 $\pm$ 3.64} & \cellcolor{my_blue}4.00 & \cellcolor{my_blue}\textbf{1.19 $\pm$ 1.21} & \cellcolor{my_blue}0.00 & \cellcolor{my_blue}\textbf{83.36 $\pm$ 0.11}
				& \cellcolor{my_blue}\textbf{38.41 $\pm$ 3.81} & \cellcolor{my_blue}4.00 & \cellcolor{my_blue}\textbf{0.70 $\pm$ 0.89} & \cellcolor{my_blue}0.00 & \cellcolor{my_blue}\textbf{83.37 $\pm$ 0.09} \\ \cline{2-12} 
			\multicolumn{1}{||c||}{}
			& CUSUM       
				& 90.90 $\pm$ 6.13 & 4.00 & \textbf{0.32 $\pm$ 0.58} & 0.00 & 83.27 $\pm$ 0.12
				& 88.23 $\pm$ 8.97 & 3.99 $\pm$ 0.10 & \textbf{0.35 $\pm$ 0.54} & 0.01 $\pm$ 0.10 & 83.27 $\pm$ 0.08 \\ 
			\multicolumn{1}{||c||}{}
			& PageHinkley 
				& 229.91 $\pm$ 13.27 & 2.26 $\pm$ 0.98 & 1.74 $\pm$ 0.98 & 1.74 $\pm$ 0.98 & 82.88 $\pm$ 0.11
				& 198.79 $\pm$ 18.72 & 3.56 $\pm$ 0.65 & 0.44 $\pm$ 0.65 & 0.44 $\pm$ 0.65 & 82.97 $\pm$ 0.10 \\
			\multicolumn{1}{||c||}{}
			& DDM         
				& 195.73 $\pm$ 22.12 & 2.76 $\pm$ 1.01 & 2.91 $\pm$ 1.96 & 1.24 $\pm$ 1.01 & 81.78 $\pm$ 2.06
				& 192.99 $\pm$ 23.82 & 2.78 $\pm$ 1.00 & 2.41 $\pm$ 1.44 & 1.22 $\pm$ 1.00 & 80.28 $\pm$ 4.11 \\
			\multicolumn{1}{||c||}{}
			& EDDM        
				& 248.46 $\pm$ 7.69 & 0.05 $\pm$ 0.22 & 21.51 $\pm$ 7.70 & 3.95 $\pm$ 0.22 & 80.65 $\pm$ 0.82
				& 247.47 $\pm$ 8.60 & 0.11 $\pm$ 0.31 & 20.22 $\pm$ 7.66 & 3.89 $\pm$ 0.31 & 80.30 $\pm$ 2.32 \\ 
			\multicolumn{1}{||c||}{}
			& RDDM        
				& 106.68 $\pm$ 11.26 & 3.99 $\pm$ 0.10 & 3.49 $\pm$ 2.47 & 0.01 $\pm$ 0.10 & 83.16 $\pm$ 0.12
				& 104.97 $\pm$ 12.06 & 3.99 $\pm$ 0.10 & 1.86 $\pm$ 1.65 & 0.01 $\pm$ 0.10 & 83.24 $\pm$ 0.09 \\
			\multicolumn{1}{||c||}{}
			& ADWIN       
				& 64.72 $\pm$ 2.79 & 4.00 $\pm$ 0.00 & 4.84 $\pm$ 2.44 & 0.00 & 83.25 $\pm$ 0.12
				& 64.48 $\pm$ 1.90 & 4.00 $\pm$ 0.00 & 3.47 $\pm$ 1.42 & 0.00 & 83.28 $\pm$ 0.08 \\
			\multicolumn{1}{||c||}{}
			& SeqDrift2   
				& 200.00 & 4.00 & 4.98 $\pm$ 1.20 & 0.00 & 82.91 $\pm$ 0.11
				& 200.00 & 4.00 & 4.39 $\pm$ 0.79 & 0.00 & 82.91 $\pm$ 0.08 \\
			\multicolumn{1}{||c||}{}
			& HDDMA-test  
				& 69.42 $\pm$ 15.51 & 4.00 & \textbf{1.28 $\pm$ 1.09} & 0.00 & \textbf{83.31 $\pm$ 0.11}
				& 83.71 $\pm$ 19.46 & 3.96 $\pm$ 0.20 & \textbf{0.48 $\pm$ 0.64} & 0.04 $\pm$ 0.20 & 83.28 $\pm$ 0.09 \\ 
			\multicolumn{1}{||c||}{}
			& HDDMW-test  
				& \textbf{35.56 $\pm$ 3.50} & 4.00 & 3.23 $\pm$ 1.95 & 0.00 & 83.27 $\pm$ 0.12
				& \textbf{35.75 $\pm$ 3.94} & 4.00 & 1.77 $\pm$ 1.39 & 0.00 & \textbf{83.36 $\pm$ 0.09} \\
			\multicolumn{1}{||c||}{}
			& FHDDM       
				& 40.55 $\pm$ 3.70 & 4.00 & \textbf{0.65 $\pm$ 0.94} & 0.00 & \textbf{83.39 $\pm$ 0.10}
				& 40.56 $\pm$ 3.72 & 4.00 & \textbf{0.25 $\pm$ 0.48} & 0.00 & \textbf{83.38 $\pm$ 0.08} \\  \hline													
		\end{tabu}

	\end{center}
\end{table*}

\begin{table*}[ht]

	\begin{center}
		
		\caption{\small Hoeffding Tree and Naive Bayes with Drift Detectors against Synthetic Data Streams with Gradual Change ($\zeta = 500$)}
		\label{tab_exp_syn_2}
		\def\arraystretch{1.01}
		\setlength\tabcolsep{1.5pt}
		\scriptsize
		
		\begin{tabu}{cr|c|c|c|c|c||c|c|c|c|c|}
			\cline{3-12}
			&             & \multicolumn{5}{c||}{Hoeffding Tree (HT)}    & \multicolumn{5}{c|}{Naive Bayes (NB)}    \\ \cline{2-12}
			\multicolumn{1}{c|}{}                         & Detector    & Delay & TP & FP & FN & Accuracy & Delay & TP & FP & FN & Accuracy \\ \hline
			
			
			\multicolumn{1}{||c||}{\multirow{13}{*}{\rotatebox[]{90}{\textsc{\textbf{Circles}} - Gradual}}}
			
			& \cellcolor{my_blue}MDDM-A
			& \cellcolor{my_blue}\textbf{71.98 $\pm$ 22.19} & \cellcolor{my_blue}3.00 & \cellcolor{my_blue}\textbf{0.27 $\pm$ 0.51} & \cellcolor{my_blue}0.00 & \cellcolor{my_blue}\textbf{86.58 $\pm$ 0.16} 
			& \cellcolor{my_blue}\textbf{161.25 $\pm$ 87.26} & \cellcolor{my_blue}2.95 $\pm$ 0.22 & \cellcolor{my_blue}\textbf{0.63 $\pm$ 0.70} & \cellcolor{my_blue}0.05 $\pm$ 0.22 & \cellcolor{my_blue}\textbf{84.14 $\pm$ 0.12} \\
			\multicolumn{1}{||c||}{}                        
			& \cellcolor{my_blue}MDDM-G
			& \cellcolor{my_blue}\textbf{69.42 $\pm$ 22.09} & \cellcolor{my_blue}3.00 & \cellcolor{my_blue}\textbf{0.36 $\pm$ 0.61} & \cellcolor{my_blue}0.00 & \cellcolor{my_blue}\textbf{86.58 $\pm$ 0.17}
			& \cellcolor{my_blue}\textbf{161.73 $\pm$ 89.49} & \cellcolor{my_blue}2.94 $\pm$ 0.24 & \cellcolor{my_blue}\textbf{0.80 $\pm$ 0.73} & \cellcolor{my_blue}0.06 $\pm$ 0.24 & \cellcolor{my_blue}\textbf{84.14 $\pm$ 0.12} \\ 
			\multicolumn{1}{||c||}{}                      
			& \cellcolor{my_blue}MDDM-E
			& \cellcolor{my_blue}\textbf{69.52 $\pm$ 22.12} & \cellcolor{my_blue}3.00 & \cellcolor{my_blue}\textbf{0.37 $\pm$ 0.61} & \cellcolor{my_blue}0.00 & \cellcolor{my_blue}\textbf{86.57 $\pm$ 0.17}
			& \cellcolor{my_blue}\textbf{161.74 $\pm$ 89.49} & \cellcolor{my_blue}2.94 $\pm$ 0.24 & \cellcolor{my_blue}\textbf{0.81 $\pm$ 0.73} & \cellcolor{my_blue}0.06 $\pm$ 0.24 & \cellcolor{my_blue}\textbf{84.14 $\pm$ 0.12} \\ 
			 \cline{2-12} 
			\multicolumn{1}{||c||}{}
			& CUSUM       
			& 220.07 $\pm$ 31.79 & 2.99 $\pm$ 0.10 & \textbf{0.04 $\pm$ 0.20} & 0.01 $\pm$ 0.10 & 86.51 $\pm$ 0.13
			& 299.78 $\pm$ 52.29 & 3.00 & \textbf{0.40 $\pm$ 0.62} & 0.00 & 84.08 $\pm$ 0.12 \\
			\multicolumn{1}{||c||}{}
			& PageHinkley 
			& 855.37 $\pm$ 56.27 & 1.79 $\pm$ 0.45 & 1.24 $\pm$ 0.47 & 1.21 $\pm$ 0.45 & 85.96 $\pm$ 0.15
			& 677.32 $\pm$ 76.30 & 2.11 $\pm$ 0.55 & 0.93 $\pm$ 0.53 & 0.89 $\pm$ 0.55 & 83.94 $\pm$ 0.13 \\
			\multicolumn{1}{||c||}{}
			& DDM         
			& 487.97 $\pm$ 82.24 & 2.78 $\pm$ 0.52 & 1.41 $\pm$ 1.24 & 0.22 $\pm$ 0.52 & 86.21 $\pm$ 0.47
			& 703.59 $\pm$ 122.67 & 1.92 $\pm$ 0.72 & 2.33 $\pm$ 1.49 & 1.08 $\pm$ 0.72 & 83.18 $\pm$ 1.61 \\ 
			\multicolumn{1}{||c||}{}
			& EDDM        
			& 987.61 $\pm$ 54.35 & 0.07 $\pm$ 0.26 & 24.61 $\pm$ 14.48 & 2.93 $\pm$ 0.26 & 84.89 $\pm$ 0.29
			& 938.27 $\pm$ 106.60 & 0.35 $\pm$ 0.50 & 31.09 $\pm$ 18.14 & 2.65 $\pm$ 0.50 & 83.12 $\pm$ 0.40 \\ 
			\multicolumn{1}{||c||}{}
			& RDDM        
			& 293.80 $\pm$ 38.52 & 2.98 $\pm$ 0.14 & 0.79 $\pm$ 1.25 & 0.02 $\pm$ 0.14 & 86.46 $\pm$ 0.16
			& 406.50 $\pm$ 69.40 & 2.99 $\pm$ 0.10 & 2.15 $\pm$ 1.94 & 0.01 $\pm$ 0.10 & 84.05 $\pm$ 0.11 \\ 
			\multicolumn{1}{||c||}{}
			& ADWIN       
			& 236.48 $\pm$ 130.94 & 2.67 $\pm$ 0.47 & 9.74 $\pm$ 3.05 & 0.33 $\pm$ 0.47 & 85.62 $\pm$ 0.19
			& 222.61 $\pm$ 57.00 & 2.99 $\pm$ 0.10 & 5.56 $\pm$ 2.57 & 0.01 $\pm$ 0.10 & \textbf{84.12 $\pm$ 0.11} \\ 
			\multicolumn{1}{||c||}{}
			& SeqDrift2
			& 202.67 $\pm$ 16.11 & 3.00 & 3.08 $\pm$ 0.90 & 0.00 & 86.47 $\pm$ 0.14
			& 276.67 $\pm$ 91.10 & 2.92 $\pm$ 0.27 & 2.49 $\pm$ 0.97 & 0.08 $\pm$ 0.27 & \textbf{84.13 $\pm$ 0.14} \\ 
			\multicolumn{1}{||c||}{}
			& HDDMA-test
			& 111.96 $\pm$ 68.22 & 2.96 $\pm$ 0.20 & 0.65 $\pm$ 0.92 & 0.04 $\pm$ 0.20 & 86.52 $\pm$ 0.20
			& 306.91 $\pm$ 107.78 & 2.91 $\pm$ 0.29 & \textbf{0.49 $\pm$ 0.69} & 0.09 $\pm$ 0.29 & 84.09 $\pm$ 0.12 \\ 
			\multicolumn{1}{||c||}{}
			& HDDMW-test
			& 94.03 $\pm$ 57.61 & 2.98 $\pm$ 0.14 & 0.73 $\pm$ 0.87 & 0.02 $\pm$ 0.14 & 86.53 $\pm$ 0.18
			& 242.43 $\pm$ 134.19 & 2.73 $\pm$ 0.44 & 1.59 $\pm$ 1.00 & 0.27 $\pm$ 0.44 & 84.11 $\pm$ 0.13 \\ 
			\multicolumn{1}{||c||}{}
			& FHDDM
			& 79.28 $\pm$ 20.64 & 3.00 & \textbf{0.17 $\pm$ 0.40} & 0.00 & \textbf{86.58 $\pm$ 0.13}
			& 166.13 $\pm$ 83.84 & 2.96 $\pm$ 0.20 & \textbf{0.43 $\pm$ 0.60} & 0.04 $\pm$ 0.20 & \textbf{84.14 $\pm$ 0.13} \\ 
			
			\hline
			\multicolumn{9}{c}{} \\ [-0.6em]
			\hline
			
			
			\multicolumn{1}{||c||}{\multirow{12}{*}{\rotatebox[]{90}{\textsc{\textbf{LED\textsubscript{0.3.1.3}}} - Gradual}}} 
			& \cellcolor{my_blue}MDDM-A      
			& \cellcolor{my_blue}\textbf{210.31 $\pm$ 73.05} & \cellcolor{my_blue}2.98 $\pm$ 0.14 & \cellcolor{my_blue}\textbf{0.03 $\pm$ 0.17} & \cellcolor{my_blue}0.02 $\pm$ 0.14 & \cellcolor{my_blue}\textbf{89.56 $\pm$ 0.04}
			& \cellcolor{my_blue}\textbf{210.31 $\pm$ 73.05} & \cellcolor{my_blue}2.98 $\pm$ 0.14 & \cellcolor{my_blue}\textbf{0.03 $\pm$ 0.17} & \cellcolor{my_blue}0.02 $\pm$ 0.14 & \cellcolor{my_blue}\textbf{89.57 $\pm$ 0.04} \\
			\multicolumn{1}{||c||}{}
			& \cellcolor{my_blue}MDDM-G      
			& \cellcolor{my_blue}\textbf{208.65 $\pm$ 73.05} & \cellcolor{my_blue}2.98 $\pm$ 0.14 & \cellcolor{my_blue}\textbf{0.03 $\pm$ 0.17} & \cellcolor{my_blue}0.02 $\pm$ 0.14 & \cellcolor{my_blue}\textbf{89.56 $\pm$ 0.04}
			& \cellcolor{my_blue}\textbf{208.65 $\pm$ 73.05} & \cellcolor{my_blue}2.98 $\pm$ 0.14 & \cellcolor{my_blue}\textbf{0.03 $\pm$ 0.17} & \cellcolor{my_blue}0.02 $\pm$ 0.14 & \cellcolor{my_blue}\textbf{89.57 $\pm$ 0.04} \\ 
			\multicolumn{1}{||c||}{}
			& \cellcolor{my_blue}MDDM-E      
			& \cellcolor{my_blue}\textbf{208.61 $\pm$ 73.05} & \cellcolor{my_blue}2.98 $\pm$ 0.14 & \cellcolor{my_blue}\textbf{0.03 $\pm$ 0.17} & \cellcolor{my_blue}0.02 $\pm$ 0.14 & \cellcolor{my_blue}\textbf{89.56 $\pm$ 0.04}
			& \cellcolor{my_blue}\textbf{208.61 $\pm$ 73.05} & \cellcolor{my_blue}2.98 $\pm$ 0.14 & \cellcolor{my_blue}\textbf{0.03 $\pm$ 0.17} & \cellcolor{my_blue}0.02 $\pm$ 0.14 & \cellcolor{my_blue}\textbf{89.57 $\pm$ 0.04} \\ 
			 \cline{2-12} 
			\multicolumn{1}{||c||}{}
			& CUSUM       
			& 300.68 $\pm$ 50.30 & 3.00 & \textbf{0.00} & 0.00 & \textbf{89.56 $\pm$ 0.03}
			& 300.61 $\pm$ 50.30 & 3.00 & \textbf{0.00} & 0.00 & \textbf{89.57 $\pm$ 0.03} \\
			\multicolumn{1}{||c||}{}
			& PageHinkley 
			& 560.30 $\pm$ 79.43 & 2.95 $\pm$ 0.26 & \textbf{0.04 $\pm$ 0.24} & 0.05 $\pm$ 0.26 & 89.35 $\pm$ 0.04
			& 559.27 $\pm$ 78.99 & 2.95 $\pm$ 0.26 & \textbf{0.04 $\pm$ 0.24} & 0.05 $\pm$ 0.26 & 89.36 $\pm$ 0.04 \\ 
			\multicolumn{1}{||c||}{}
			& DDM         
			& 444.13 $\pm$ 79.82 & 2.97 $\pm$ 0.17 & 0.32 $\pm$ 0.58 & 0.03 $\pm$ 0.17 & 89.47 $\pm$ 0.56
			& 446.23 $\pm$ 82.12 & 2.96 $\pm$ 0.20 & 0.33 $\pm$ 0.58 & 0.04 $\pm$ 0.20 & 89.29 $\pm$ 1.15 \\ 
			\multicolumn{1}{||c||}{}
			& EDDM        
			& 954.97 $\pm$ 62.98 & 0.66 $\pm$ 0.71 & 5.97 $\pm$ 1.69 & 2.34 $\pm$ 0.71 & 88.33 $\pm$ 0.50
			& 949.61 $\pm$ 68.94 & 0.70 $\pm$ 0.73 & 6.33 $\pm$ 1.96 & 2.30 $\pm$ 0.73 & 88.32 $\pm$ 0.53 \\ 
			\multicolumn{1}{||c||}{}
			& RDDM        
			& 321.88 $\pm$ 50.94 & 2.98 $\pm$ 0.14 & 0.61 $\pm$ 0.96 & 0.02 $\pm$ 0.14 & \textbf{89.63 $\pm$ 0.04}
			& 321.80 $\pm$ 50.94 & 2.98 $\pm$ 0.14 & 0.61 $\pm$ 0.96 & 0.02 $\pm$ 0.14 & \textbf{89.63 $\pm$ 0.04} \\ 
			\multicolumn{1}{||c||}{}
			& ADWIN       
			& 521.47 $\pm$ 239.71 & 2.40 $\pm$ 0.77 & 474.95 $\pm$ 14.12 & 0.60 $\pm$ 0.77 & 72.25 $\pm$ 0.49
			& 521.36 $\pm$ 238.56 & 2.36 $\pm$ 0.76 & 465.41 $\pm$ 12.53 & 0.64 $\pm$ 0.76 & 72.73 $\pm$ 0.44 \\
			\multicolumn{1}{||c||}{}
			& SeqDrift2   
			& 426.00 $\pm$ 173.31 & 2.78 $\pm$ 0.44 & 277.0 $\pm$ 47.5 & 0.22 $\pm$ 0.44 & 76.51 $\pm$ 2.28
			& 445.33 $\pm$ 192.27 & 2.75 $\pm$ 0.46 & 278.8 $\pm$ 47.5 & 0.25 $\pm$ 0.46 & 76.54 $\pm$ 2.25 \\ 
			\multicolumn{1}{||c||}{}
			& HDDMA-test  
			& 295.03 $\pm$ 85.29 & 2.98 $\pm$ 0.20 & 0.16 $\pm$ 0.44 & 0.02 $\pm$ 0.20 & \textbf{89.58 $\pm$ 0.05}
			& 295.85 $\pm$ 83.23 & 2.98 $\pm$ 0.20 & 0.17 $\pm$ 0.47 & 0.02 $\pm$ 0.20 & \textbf{89.58 $\pm$ 0.05} \\ 
			\multicolumn{1}{||c||}{}
			& HDDMW-test  
			& 259.18 $\pm$ 87.25 & 2.95 $\pm$ 0.26 & 0.08 $\pm$ 0.31 & 0.05 $\pm$ 0.26 & \textbf{89.56 $\pm$ 0.04}
			& 259.17 $\pm$ 87.21 & 2.95 $\pm$ 0.26 & \textbf{0.05 $\pm$ 0.22} & 0.05 $\pm$ 0.26 & \textbf{89.56 $\pm$ 0.03} \\
			\multicolumn{1}{||c||}{}
			& FHDDM       
			& 220.40 $\pm$ 76.00 & 2.97 $\pm$ 0.22 & \textbf{0.03 $\pm$ 0.22} & 0.03 $\pm$ 0.22 & \textbf{89.56 $\pm$ 0.04}
			& 220.40 $\pm$ 76.00 & 2.97 $\pm$ 0.22 & \textbf{0.03 $\pm$ 0.22} & 0.03 $\pm$ 0.22 & \textbf{89.57 $\pm$ 0.04} \\ \hline													
		\end{tabu}
		
	\end{center}
\end{table*}

\noindent \textbf{Discussion I - \textsc{Sine1} and \textsc{Mixed} (Abrupt Drift):}	As represented in Table \ref{tab_exp_syn_1}, MDDMs and HDDM\textsubscript{W-test} detected concept drifts with shorter delays against \textsc{Sine1} and \textsc{Mixed} data streams. MDDM, FHDDM, CUSUM and HDDM\textsubscript{A-test} resulted in the lowest false positive rates. This observation may indicate that MDDM, FHDDM, CUSUM, and HDDM\textsubscript{A-test} are more accurate.
Although RDDM had shorter detection delays and false negative rates compared to DDM and EDDM, it caused higher false positive rates.
EDDM had the highest false positive rates. Moreover, EDDM had the highest false negative rates since it could not detect concept drifts within the acceptable delay length.
MDDMs showed comparable results against the other methods.
As shown in Table \ref{tab_exp_syn_1}, for the Hoeffding Tree classifier, the highest classification accuracies was obtained with MDDMs and FHDDM, since they detected drifts with the shortest delays and the lowest false positive rates. Similar observations apply to the Naive Bayes classifier.
It may be noticed that the false positive rate is lower for Naive Bayes.
This suggests that the Naive Bayes classifier represented the decision boundaries more accurately for noisy \textsc{Sine1} and \textsc{Mixed} data streams.

\noindent \textbf{Discussion II - \textsc{Circles} and \textsc{LED} (Gradual Drift):}
Table \ref{tab_exp_syn_2} shows the results with the Hoeffding Tree and Naive Bayes classifiers against the \textsc{Circles} and \textsc{LED} data streams.
The MDDM algorithms resulted in the shortest concept drift detection delays, followed by FHDDM and HDDM\textsubscript{W-test}.
Compared to FHDDM, MDDMs detected concept drifts faster because of its weighting schemes which favor the most recent elements.
On the other hand, EDDM produced the longest drift detection delays. It also had the highest false negative rates.
MDDMs, FHDDM, CUSUM, RDDM, and HDDMs had the highest true positive rates.
EDDM showed the highest false positive rates against the \textsc{Circles} data streams.
We achieved higher accuracies with Hoeffding Tree than with Naive Bayes against the \textsc{Circles} data stream.
In the case of the \textsc{LED} data streams, ADWIN and SeqDrift2 triggered a relatively large number of false alarms. Indeed, this could be potentially alleviated by decreasing their confidence levels, i.e.\ $\delta$, to make their tests more restrictive.
SeqDrift2 caused fewer false positives compared to ADWIN, since it applies a more conservative test (Bernstein's inequality).
Although RDDM outperformed DDM in all cases, in terms of detection delays and false negative rates, it showed higher false positive rates.
Finally, MDDMs, FHDDM, and HDDMs led to the highest accuracies with both classifiers.

\noindent \textbf{Discussion III - MDDM Variants: }
Frequently, MDDM-G and MDDM-E have shorter drift detection delays than MDDM-A. The reason is to be found in the fact that they both utilize an exponential weighting scheme (i.e. more weight is put on the most recent entries which are the ones required for faster detection) as opposed to MDDM-A which has a linear one.
The reader will notice that the false positive rates of these two variants against the two streams with gradual change, namely \textsc{Circles} and \textsc{LED}, were higher than those of MDDM-A. This is a consequence of the fact that MDDM-A put more emphasis on the older entries in the window, which, in these cases are beneficial to the learning process.  All three variants had comparable levels of accuracy.
In general, one may observe that an exponential-like scheme is beneficial in scenarios when faster detection is required.
It follows that the optimal shape for the weighting function is data, context and application dependent.

\subsubsection{Real-world Data Streams}

There is a consensus among researchers that the locations and/or the presence of concept drift in the \textsc{Electricity}, \textsc{Forest Covertype}, and \textsc{Poker hand} data streams are not known \cite{frias2015online, pesaranghader2016fast, huang2015drift, bifet2009new, losing2018incremental}. This implies, in turn, that the drift detection delay as well as the false positive and false negative rates cannot be determined since the knowledge of the drift locations is necessary in order to evaluate these quantities. Consequently, our evaluation is based on the overall accuracy and the number of alarms for concept drifts issued by each drift detector. We have also considered \textit{blind adaptation} and \textit{no detection} approaches as benchmarks for our experiments. In the \textit{blind adaptation}, the classifier is retrained ab initio at every $100$ instances. The classifiers are trained without drift detectors in the case of \textit{no detection}. Similar to \cite{pesaranghader2016fast}, a window of size $25$ was selected for FHDDM and MDDMs against real-world datasets. Our experiments have shown that this choice is optimal in terms of accuracy.

\begin{table*}[hpb]
	\begin{center}
		\caption{\small Hoeffding Tree (HT) and Naive Bayes (NB) against Real-world Data Stream}
		\label{table_ht_nb_real}
		\def\arraystretch{1.01}
		\setlength\tabcolsep{1.5pt}
		\scriptsize
		\begin{tabu}{r|C{1cm}|C{1cm}|C{1cm}|C{1cm}|C{1cm}||C{1cm}|C{1cm}|C{1cm}|C{1cm}||C{1cm}|C{1cm}|C{1cm}|C{1cm}|}
			\tabucline{3-14}
			\multicolumn{1}{c}{} & & \multicolumn{4}{c||}{\textsc{Electricity}} & \multicolumn{4}{c||}{\textsc{Forest Covertype}} & \multicolumn{4}{c|}{\textsc{Poker hand}} \\ \tabucline{3-14} 
			\multicolumn{1}{c}{} & & \multicolumn{2}{c|}{HT} & \multicolumn{2}{c||}{NB} & \multicolumn{2}{c|}{HT} & \multicolumn{2}{c||}{NB} & \multicolumn{2}{c|}{HT} & \multicolumn{2}{c|}{NB} \\ \tabucline[2pt]{2-14}
			\multicolumn{1}{c|}{} & \multicolumn{1}{c||}{\cellcolor{whitesmoke}Detector} & \cellcolor{whitesmoke}Alarms & \cellcolor{whitesmoke}Acc. & \cellcolor{whitesmoke}Alarms & \cellcolor{whitesmoke}Acc. & \cellcolor{whitesmoke}Alarms & \cellcolor{whitesmoke}Acc. & \cellcolor{whitesmoke}Alarms & \cellcolor{whitesmoke}Acc. & \cellcolor{whitesmoke}Alarms & \cellcolor{whitesmoke}Acc. & \cellcolor{whitesmoke}Alarms & \cellcolor{whitesmoke}Acc. \\ \tabucline[1pt]{2-14}
			& \multicolumn{1}{r||}{\cellcolor{my_blue}MDDM-A}   		& \cellcolor{my_blue}105 & \cellcolor{my_blue}84.60 & \cellcolor{my_blue}126 & \cellcolor{my_blue}83.47 & \cellcolor{my_blue}1963 & \cellcolor{my_blue}85.33 & \cellcolor{my_blue}2022 & \cellcolor{my_blue}85.38 & \cellcolor{my_blue}2036 & \cellcolor{my_blue}76.89 & \cellcolor{my_blue}2145 & \cellcolor{my_blue}76.83 \\
			& \multicolumn{1}{r||}{\cellcolor{my_blue}MDDM-G}   		& \cellcolor{my_blue}105 & \cellcolor{my_blue}84.60 & \cellcolor{my_blue}126 & \cellcolor{my_blue}83.47 & \cellcolor{my_blue}1966 & \cellcolor{my_blue}85.35 & \cellcolor{my_blue}2025 & \cellcolor{my_blue}85.39 & \cellcolor{my_blue}2034 & \cellcolor{my_blue}76.89 & \cellcolor{my_blue}2149 & \cellcolor{my_blue}76.83 \\
			& \multicolumn{1}{r||}{\cellcolor{my_blue}MDDM-E}   		& \cellcolor{my_blue}105 & \cellcolor{my_blue}84.60 & \cellcolor{my_blue}126 & \cellcolor{my_blue}83.47 & \cellcolor{my_blue}1966 & \cellcolor{my_blue}85.35 & \cellcolor{my_blue}2025 & \cellcolor{my_blue}85.39 & \cellcolor{my_blue}2034 & \cellcolor{my_blue}76.89 & \cellcolor{my_blue}2149 & \cellcolor{my_blue}76.83 \\
			\tabucline{2-14}
			& \multicolumn{1}{r||}{CUSUM}     		& 22 & 81.71 & 28 & 79.21 & 226 & 83.01 & 286 & 81.55 & 617 & 72.85 & 659 & 72.54 \\
			& \multicolumn{1}{r||}{PageHinkley}	& 6 & 81.95 & 10 & 78.04 & 90 & 81.65 & 117 & 80.06 & 403 & 71.30 & 489 & 70.67 \\
			& \multicolumn{1}{r||}{DDM}       		& 169 & \textbf{85.41} & 143 & 81.18 & 4301 & \textbf{87.35} & 4634 & \textbf{88.03} & 1046 & 72.74 & 433 & 61.97 \\
			& \multicolumn{1}{r||}{EDDM}      		& 191 & 84.91 & 203 & \textbf{84.83} & 2466 & 86.00 & 2416 & 86.08 & 4806 & \textbf{77.30} & 4863 & \textbf{77.48} \\
			& \multicolumn{1}{r||}{RDDM}      		& 143 & 85.18 & 164 & 84.19 & 2671 & 86.42 & 2733 & 86.86 & 2512 & 76.70 & 2579 & 76.67 \\
			& \multicolumn{1}{r||}{ADWIN}     		& 65 & 83.23 & 88 & 81.03 & 1062 & 83.36 & 1151 & 83.24 & 1358 & 73.84 & 1388 & 73.69 \\
			& \multicolumn{1}{r||}{SeqDrift2} 		& 59 & 82.83 & 60 & 79.68 & 710 & 82.85 & 757 & 82.44 & 1322 & 72.51 & 1395 & 72.25 \\
			& \multicolumn{1}{r||}{HDDM\textsubscript{A-test}}     & 210 & \textbf{85.71} & 211 & \textbf{84.92} & 3695 & 87.24 & 3284 & 87.42 & 2565 & 76.40 & 2615 & 76.48 \\
			& \multicolumn{1}{r||}{HDDM\textsubscript{W-test}}     & 117 & 85.06 & 132 & 84.09 & 2342 & 85.97 & 2383 & 86.22 & 2211 & 77.11 & 2312 & 77.11 \\
			& \multicolumn{1}{r||}{FHDDM}     		& 90 & 84.59 & 109 & 83.13 & 1794 & 85.08 & 185 & 85.09 & 1876 & 76.72 & 1928 & 76.68 \\
			\tabucline{2-14}
			
			\multicolumn{14}{c}{}
			\\ [-0.875em] \tabucline{2-14}
			
			& \multicolumn{1}{r||}{\cellcolor{my_amber}Blind\textsubscript{$|\mbox{\textsc{w}}|=100$}}     & \cellcolor{my_amber}453 & \cellcolor{my_amber}84.26 & \cellcolor{my_amber}453 & \cellcolor{my_amber}84.82 & \cellcolor{my_amber}5810 & \cellcolor{my_amber}\textbf{87.24} & \cellcolor{my_amber}5810 & \cellcolor{my_amber}\textbf{87.70} & \cellcolor{my_amber}8292 & \cellcolor{my_amber}\textbf{77.96} & \cellcolor{my_amber}8292 & \cellcolor{my_amber}\textbf{78.18} \\
			& \multicolumn{1}{r||}{No Detection}        & --- & 79.20 & --- & 73.36 & --- & 80.31 & --- & 60.52 & --- & 76.07 & --- & 59.55 \\ \tabucline{2-14}
			
			\multicolumn{14}{c}{}
			\\ [-0.5em] \tabucline[2pt]{1-14}
			
			\multicolumn{1}{|c|}{\multirow{3}{*}{\rotatebox[]{90}{\makecell{$(\delta_w)$ \\ $0.001$}}}} 
				& \multicolumn{1}{r||}{\cellcolor{my_green}MDDM-A} 
				& \cellcolor{my_green}180 & \cellcolor{my_green}\textbf{85.79} & \cellcolor{my_green}208 & \cellcolor{my_green}\textbf{85.00} & \cellcolor{my_green}3253 & \cellcolor{my_green}\textbf{87.03} & \cellcolor{my_green}3221 & \cellcolor{my_green}\textbf{87.27} 
				& \cellcolor{my_green}4320 & \cellcolor{my_green}\textbf{77.82} & \cellcolor{my_green}4378 & \cellcolor{my_green}\textbf{78.03} \\
			\multicolumn{1}{|c|}{} 
				& \multicolumn{1}{r||}{\cellcolor{my_green}MDDM-G} 
				& \cellcolor{my_green}182 & \cellcolor{my_green}\textbf{85.78} & \cellcolor{my_green}209 & \cellcolor{my_green}\textbf{85.01} & \cellcolor{my_green}3270 & \cellcolor{my_green}\textbf{87.05} & \cellcolor{my_green}3231 & \cellcolor{my_green}\textbf{87.29} 
				& \cellcolor{my_green}4370 & \cellcolor{my_green}\textbf{77.83} & \cellcolor{my_green}4425 & \cellcolor{my_green}\textbf{78.05} \\
			\multicolumn{1}{|c|}{} 
				& \multicolumn{1}{r||}{\cellcolor{my_green}MDDM-E}		
				& \cellcolor{my_green}182 & \cellcolor{my_green}\textbf{85.78} & \cellcolor{my_green}209 & \cellcolor{my_green}\textbf{85.01} & \cellcolor{my_green}3270 & \cellcolor{my_green}\textbf{87.06} & \cellcolor{my_green}3234 & \cellcolor{my_green}\textbf{87.29} 
				& \cellcolor{my_green}4369 & \cellcolor{my_green}\textbf{77.83} & \cellcolor{my_green}4427 & \cellcolor{my_green}\textbf{78.06} \\
			
			\hline \hline
			
			\multicolumn{1}{|c|}{\multirow{3}{*}{\rotatebox[]{90}{\makecell{($\delta_w$) \\ $0.01$}}}}
				& \multicolumn{1}{r||}{\cellcolor{my_green}MDDM-A} 
				& \cellcolor{my_green}256 & \cellcolor{my_green}\textbf{85.86} & \cellcolor{my_green}265 & \cellcolor{my_green}\textbf{85.60} 
				& \cellcolor{my_green}3884 & \cellcolor{my_green}\textbf{87.63} & \cellcolor{my_green}3791 & \cellcolor{my_green}\textbf{87.95} 
				& \cellcolor{my_green}6075 & \cellcolor{my_green}\textbf{78.18} & \cellcolor{my_green}6099 & \cellcolor{my_green}\textbf{78.51} \\
			\multicolumn{1}{|c|}{} 
				& \multicolumn{1}{r||}{\cellcolor{my_green}MDDM-G}
				& \cellcolor{my_green}252 & \cellcolor{my_green}\textbf{85.98} & \cellcolor{my_green}265 & \cellcolor{my_green}\textbf{85.78} 
				& \cellcolor{my_green}3969 & \cellcolor{my_green}\textbf{87.73} & \cellcolor{my_green}3856 & \cellcolor{my_green}\textbf{88.05} 
				& \cellcolor{my_green}6095 & \cellcolor{my_green}\textbf{78.21} & \cellcolor{my_green}6118 & \cellcolor{my_green}\textbf{78.54} \\
			\multicolumn{1}{|c|}{} 
				& \multicolumn{1}{r||}{\cellcolor{my_green}MDDM-E} 
				& \cellcolor{my_green}252 & \cellcolor{my_green}\textbf{85.97} & \cellcolor{my_green}266 & \cellcolor{my_green}\textbf{85.77} 
				& \cellcolor{my_green}3980 & \cellcolor{my_green}\textbf{87.73} & \cellcolor{my_green}3864 & \cellcolor{my_green}\textbf{88.05} 
				& \cellcolor{my_green}6091 & \cellcolor{my_green}\textbf{78.21} & \cellcolor{my_green}6116 & \cellcolor{my_green}\textbf{78.54} \\
			\tabucline[2pt]{1-14}
		\end{tabu}
	\end{center}
\end{table*}

Table \ref{table_ht_nb_real} presents the experimental results for \textsc{Electricity}, \textsc{Forest Covertype}, and \textsc{Poker hand} data streams with the Hoeffding Tree (HT) and Naive Bayes (NB) classifiers.
Firstly, the Hoeffding Tree classifier showed higher classification accuracies compared to Naive Bayes when executed without drift detector. This suggests that the Hoeffding Tree classifier could branch out and adequately reflect the new patterns.
Secondly, both classifiers achieved higher classification accuracies by using drift detection.
Although this observation indicates that using drift detection methods is beneficial compared to the \textit{no detection} case, it does not necessarily mean that a drift detector outperforms the others.
Indeed, in a recent study by Bifet et al.\ \cite{bifet2017classifier}, it was found that blind detection has the highest classification accuracies, against the \textsc{Electricity} and \textsc{Forest Covertype} data streams. Based on multiple experiments, Bifet et al.\ \cite{bifet2017classifier} concluded that this behavior may be explained by the temporal dependencies in between the instances of the streams.
As shown in Table \ref{table_ht_nb_real}, a drift detection method with a higher number of alarms usually led to a higher classification accuracy. In such a case, a classifier learns from a small portion of the data stream where almost all instances are labeled with a common label (this refers to temporal dependencies among examples as stated by Bifet et al.\ \cite{bifet2017classifier}).
To support this observation, as mentioned earlier, we considered a blind adaptation as a benchmark. As shown in the same table, the blind adaptation led to the highest or the second highest classification accuracies. We further extended our experiments by running MDDMs with higher values of $\delta_w$. Recall that a higher $\delta$ implies that the drift detection technique is less conservative. As indicated in the table, as MDDMs became less conservative, the number of alarms as well as the classification accuracies increased. Therefore, because of temporal dependencies, both classifiers repeatedly learned from instances presenting the same labels between two consecutive alarms.

In summary, we concluded that using drift detection methods against real-world data streams is beneficial. Nevertheless, we are not in a position to make a strong statement based solely on the accuracy because (1) the location of the drift is unknown, and (2) because of the temporal dependencies in between instances \cite{bifet2017classifier}. MDDM consistently led to high classification accuracies. Particularly, MDDM achieved the highest classification accuracies in all cases when the value of $\delta_w$ increased from $10^{-6}$ to $0.001$ and $0.01$.

\section{Conclusion}
\label{sec_conclusion}

Sensor networks, smart houses, intelligent transportation, autopilots are examples of technologies operating in evolving environments where experiencing concept drifts over time is commonplace.
In order for the learning process to be more accurate and efficient in evolving environments, concept drifts should be detected rapidly with false negative rate as small as possible. In this research paper, we introduced the McDiarmid Drift Detection Methods (MDDMs) for detecting concept drifts with shorter delays and lower false negative rates. We conducted various experiments to compare MDDMs against the state-of-the-art. Our experimental results indicated that MDDMs outperformed existing methods in terms of drift detection delay, false negative, false positive, and classification accuracy rates.

In this paper, we considered incremental learning against a single stream while evaluating the drift detection methods accordingly. We aim to investigate streams with heterogeneous concept drifts, i.e.\ streams in which different drift types and rates overlap. 
We are in addition interested to compare the performance of MDDMs with proactive drift detection methods, such as the DetectA algorithm \cite{escovedo2018detecta}, particularly regarding the detection delay.
Further, we aim to apply the notion of multiple sliding windows stacking, as introduced by Pesaranghader et al.\ \cite{pesaranghader2017reservoir}, to the MDDM approaches.
We also plan to investigate adaptive ensemble approaches and self-adjusting algorithms \cite{losing2016knn}. 
Finally, we will assess the performance of drift detection methods against time-series data streams.

\section*{Acknowledgment}

The authors wish to acknowledge financial support by the Canadian Natural Sciences and Engineering Research Council (NSERC) as well as the  Ontario Trillium Scholarship (OTS).

\bibliographystyle{IEEEtran}
\bibliography{mddm}

\end{document}